\documentclass[10pt,conference,a4paper]{IEEEtran}
\IEEEoverridecommandlockouts
\usepackage{subfigure}
\usepackage{cite}
\usepackage{amsmath,amssymb,amsfonts}
\usepackage{algorithmic}
\usepackage{graphicx}
\usepackage{textcomp}
\usepackage{xcolor}
\def\BibTeX{{\rm B\kern-.05em{\sc i\kern-.025em b}\kern-.08em
    T\kern-.1667em\lower.7ex\hbox{E}\kern-.125emX}}

\usepackage{url}
\usepackage{booktabs}

\usepackage{amsmath}
\usepackage{amssymb}

\usepackage{graphicx}
\usepackage[colorlinks=true,allcolors=red]{hyperref}
\usepackage{adjustbox}

\usepackage{soul}

\begin{document}

\title{Class-incremental Learning \\ with Pre-allocated Fixed Classifiers}

\author{\IEEEauthorblockN{Federico Pernici}
\IEEEauthorblockA{\textit{University of Florence}}
\and
\IEEEauthorblockN{Matteo Bruni}
\IEEEauthorblockA{\textit{University of Florence}}
\and
\IEEEauthorblockN{Claudio Baecchi}
\IEEEauthorblockA{\textit{University of Florence}\\[5pt]
\{name.surname\}@unifi.it}
\and
\IEEEauthorblockN{Francesco Turchini}
\IEEEauthorblockA{\textit{University of Florence}}
\and
\IEEEauthorblockN{Alberto Del Bimbo}
\IEEEauthorblockA{
\textit{University of Florence}}
}

\maketitle

\maketitle

\begin{abstract}
In class-incremental learning, a learning agent faces a stream of data with the goal of learning new classes while not forgetting  previous ones. Neural networks are known to suffer under this setting, as they forget previously acquired knowledge. To address this problem, effective methods exploit past data stored in an episodic memory while expanding the final classifier nodes to accommodate the new classes.

In this work, we substitute the expanding classifier with a novel fixed classifier
in which a number of pre-allocated output nodes are subject to the classification loss right from the beginning of the learning phase. Contrarily to the standard expanding classifier, this allows: (a) the output nodes of future unseen classes to firstly see negative samples since the beginning of learning together with the positive samples that incrementally arrive; (b) to learn features that do not change their geometric configuration as novel classes are incorporated in the learning model.

Experiments with public
datasets show that the proposed approach is as effective as the expanding classifier while exhibiting novel intriguing properties of the internal feature representation that are otherwise not-existent. Our ablation study on pre-allocating a large number of classes further validates the approach.
\end{abstract}


%
\IEEEpeerreviewmaketitle

\section{Introduction}

Natural intelligent systems learn incrementally by continuously receiving information over time. They learn new concepts adapting to changes in the environment by  leveraging past experiences. A remarkable capability of these systems is that learning of new concepts is achieved while \emph{not} forgetting previous ones. In contrast, current Deep Learning models, when updated with novel incoming data, suffer from \emph{catastrophic forgetting}: the tendency of Neural Networks to completely and abruptly forget previously learned information \cite{mccloskey1989catastrophic,ratcliff1990connectionist,goodfellow2013empirical}. This problem is related to the plasticity/stability dilemma in incremental learning \cite{grossberg1982does}.
Too much ``plasticity'' leads to catastrophic forgetting, too much ``stability'' leads to an inability to adapt to novel information. 
Continual Learning \cite{chen2018lifelong,parisi2019continual} specifically addresses this problem, bringing machine learning closer to natural learning. In this learning scenario, the agent is presented with a stream of tasks and each new task is learned by reusing and combining the knowledge acquired while learning previous tasks. As the learning agent is processing a stream, it cannot store all examples seen in the past. 

Continual learning has recently received increasing attention and several methods have been developed
\cite{aljundi2019interfere,rao2019continual,Hou_2019_CVPR,liu2020mnemonics,wortsman2020supermasks,liu2020incremental,kurle2019continual,oshg2019hypercl}. However, despite the intense research efforts, the gap in performance with respect to offline multi-task learning makes continual learning an open problem. Most of the techniques have focused on a sequence of tasks in which both the identity of the task (task label) and boundaries between tasks are provided \cite{kirkpatrick2017overcoming,nguyen2017variational,zenke2017continual,shin2017continual}. Thus, many of these methods fail to capture real-world continual learning, with unknown task labels \cite{farquhar2018towards} \cite{Hsu18_EvalCL}. A typical example that illustrates the difference between using or not the task labels is the Split MNIST experiment, in which the ten digits of the well known handwritten dataset are split into five classification tasks of two-class each. The model has five different final classification layers, one for each task. Those classifiers (i.e. output heads) are indexed by the task identity (1 to 5) that is given at testing time. 

\begin{figure}[t!]
    \vspace{0.6cm}
    \centering
    \includegraphics[width=1.0\columnwidth]{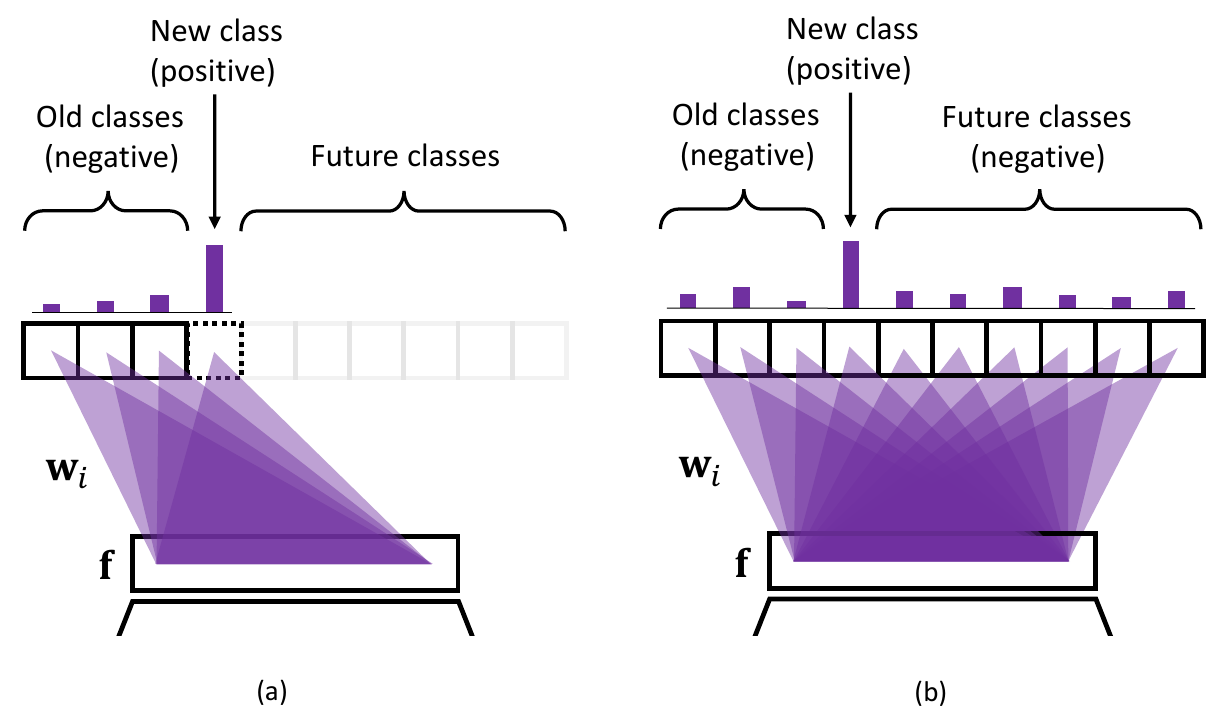}
    \caption{Class-incremental classifiers. \emph{(a)}: Expanding classifier. \emph{(b)}: Pre-allocated classifier. The latter can exploit unseen future classes as negative examples.}
    \label{fig:true_intro}
\end{figure}

This scenario is shown to be easier than class-incremental learning (CIL) since the selection of the output head is given by the task identity \cite{farquhar2018towards}. CIL is typically addressed with single-headed variants that do not require task identity, where the model always performs prediction over all classes (i.e. all digits 0 to 9)  \cite{rebuffi2017icarl,parisi2019continual,maltoni2019continuous,farquhar2018towards,Hsu18_EvalCL,vandeven2019three}.

In CIL the single head final layer of a Neural Network is \emph{expanded} with an output node when a new class arrives (multiple new classes are expanded with multiple nodes); thus, in general, during learning, an output node sees, according to the samples in the current random batch, positive and negative samples in the \emph{newly arrived} class and in the \emph{old seen classes} (i.e. the remaining), respectively (Fig.\ref{fig:true_intro}(a)).

In this paper, we address CIL using a novel classifier in which a number of pre-allocated output nodes are subject to the classification loss right from the beginning. This allows the output nodes of yet \emph{unseen classes} to firstly see negative samples since the beginning of learning together with the positive samples that incrementally arrive (Fig.\ref{fig:true_intro}(b)). Contrarily to the \emph{expanding} classifier, in our formulation, the output nodes can learn from the beginning of the learning phase. This is achieved by \emph{pre-allocating} a special classifier with a large number of output nodes in which the weights are \emph{fixed} (i.e. not undergoing learning) and set to values taken from the coordinate vertices of regular polytopes \cite{pernici2019fix}. The classification layer so defined has two intriguing properties. The first is that the features do not change their geometric configuration as novel classes are incorporated in the learning model. {The second is that a very large number of classes can be pre-allocated with no loss of accuracy. This allows the method to meet the underlying assumption of lifelong learning as for the case of the expanding classifier.
}

\vspace{0.5cm}
\section{Related Work}

\subsection{Continual Learning}

Continual learning has been extensively studied in literature \cite{chen2018lifelong,parisi2019continual}. Prior works can be broadly categorized into three main categories:
(1) regularization, (2) dynamic architecture methods and (3) episodic memory-based (also termed Experience Replay).

{\bf{Regularization}.} Regularization-based approaches reduce forgetting by restricting the updates in parameters that were important for previous tasks.
Elastic Weighted Consolidation (EWC) imposes constraints on network parameters to reproduce biological mechanism of consolidation~\cite{kirkpatrick2017overcoming}. Online-EWC~\cite{progresscompress} optimizes EWC approach for multiple tasks, overcoming the complexity of original EWC which scales linearly with number of tasks. Synaptic Intelligence (SI) also replicates biological mechanism of synapses, preventing parameters (synapses) to change based on the relevance of each parameter for the considered task~\cite{zenke2017continual}. Memory Aware Synapses (MAS) tackles the problem in a similar fashion, based on the relevance of each parameter to the task.
When the number of tasks is large, the regularization of past tasks becomes obsolete, leading to representation drift \cite{titsias2019functional}.

{\bf{Dynamic Architecture}.} 
Second, dynamic architecture or modular approaches add new modules to the model architecture as new tasks are learned.~\cite{xiao2014error} grows a network searching for similarities between known classes and unseen classes, organizing them into a hierarchy. Predictions are made by visiting the hierarchy, from the super-classes down to the specific class. \cite{cortes2017adanet} exploits boosting algorithm to control network architecture growth balancing its complexity with empirical risk minimization. The work in \cite{rusu2016progressive} proposes a network structure organized in columns. Each column is a network which learns a new task, sharing features learned by other columns via lateral connections. Sparse regularization is employed in~\cite{lee2017lifelong} to decide how many parameters add to each network layer when new tasks are learned. Then, selective retraining is performed.\\
While modular architectures overcome forgetting by design, these approaches do not scale with the number of tasks as memory requirements increase with the number of tasks.

{\bf{Experience Replay (ER)}.} 
Third, Experience Replay methods store a few examples from past tasks in an ``episodic memory'', to be revisited when training for a new task. In contrast to modular approaches, memory-based methods add a relatively small memory overhead for each new task. The concept of experience replay in class-incremental learning has been introduced in~\cite{rebuffi2017icarl}. By means of rehearsal technique, new and old data are combined when new tasks are learned in order to prevent catastrophic forgetting of old tasks. \cite{lopez2017gradient} presents Gradient Episodic Memory technique which does not store and reuse old samples but allows transfer learning between tasks by storing old gradients and updating them to prevent forgetting. An improved, memory-efficient version of GEM is obtained by considering the average of the losses of all tasks rather than each individual loss of single tasks~\cite{chaudhry2019agem}. Full data rehearsal may prevent catastrophic forgetting, but it is unfeasible due to important memory impact, so ~\cite{hayes2018memory} implements memory-efficient buffer techniques to perform rehearsal without the need for retaining all samples. \cite{isele2018selective} aims at finding data distribution which can keep optimal performance level overall tasks. This is achieved by choosing an adequate strategy to build data memory, exploiting the biological mechanic of replaying experiences. \cite{riemer2018learning} introduces a technique to avoid learning interference provoked by data coming from different source domains. Dual-memory incremental learning is exploited in~\cite{Belouadah_2019_ICCV} to keep track of statistics of past classes, in order to rebalance their prediction scores in later stages of learning.
Memory-based methods have currently shown state-of-the-art.

\subsection{Fixed Classifiers}
{ Dynamically freezing\footnote{The terms frozen and fixed are used interchangeably. }  weights is a form of implicit dynamic architecture in which some selected weights are not undergoing learning \cite{serra2018overcoming,Mallya18Piggyback,mallya2018packnet,maltoni2019continuous}. When freezing is applied to the final classification layer, class decision boundaries remain stationary during learning. This was exploited in \cite{jung2016less,jung2018less} to reduce catastrophic forgetting in domain adaptation. They show that a frozen classifier, together with a distillation loss on the features 
preserve the geometric configuration of old classes. These two constraints are recently shown to be resilient to catastrophic forgetting in class incremental learning, and in incremental few-shot learning \cite{Hou_2019_CVPR,liu2020incremental}, respectively. Freezing the classifier is also the key strategy used in \cite{Shen_2020_CVPR} to learn visual features that are compatible with features computed with different CNN models. Compatibility between features from different models learned at different times means that if such features are used to compare images, then ``new'' features can be compared directly to ``old'' features, so they can be used interchangeably. This enables visual search systems to avoid re-computing new features for all previously seen images when updating the models.
This capability may enable incremental learning of features in more realistic scenarios in which classes can be revisited
\cite{pernici2018memory},\cite{aljundi2019task},\cite{pernici2020self},\cite{caccia2020online}.

Fixing the final classification layer in multi-task supervised classification  has been explored in detail in \cite{hoffer2018fix,pernici2019fix,pernici2019maximally,qian2020we} showing that it causes little or no reduction in accuracy for common datasets, while allowing a noticeable reduction in trainable parameters.
Fixed classifiers have also an important role in theoretical convergence analysis of training neural network models with batch-norm \cite{Ioffe17}. It is shown recently in \cite{li2019exponential} that CNNs with a fixed classifier and batch-norm in each layer establish a principle of equivalence between different learning rate schedules.
}

\vspace{0.5cm}
\section{Contributions}

\begin{itemize}
   \item We introduce a novel approach for class-incremental learning that keeps features in a constant specific spatial configuration distributed at equal angles maximizing the available space. 
   \item The approach exploits negative samples from unseen classes since the beginning of learning.
   \item We achieve similar results with respect to several important baselines on standard benchmarks.
\end{itemize}

\vspace{0.5cm}
\section{Proposed Method}

\subfigcapskip = 0.1cm
\subfigbottomskip = 0.3cm
\begin{figure}[t!]
    \vspace{0.8cm}
    \centering
    \subfigure[]{
    \includegraphics[width=1.0\columnwidth]{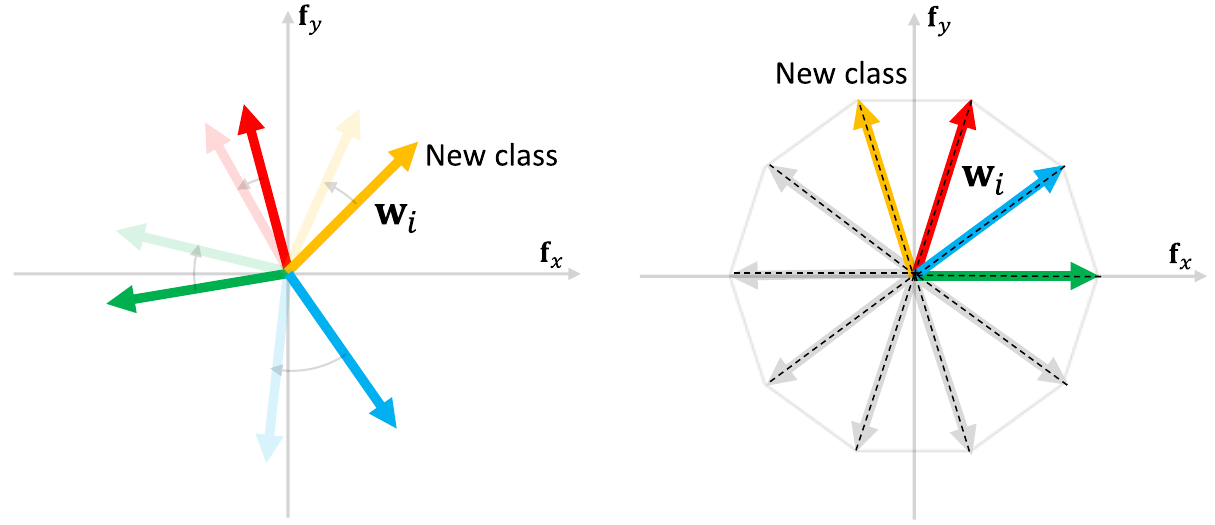}
    }
    \subfigure[]{
    \includegraphics[width=1.0\columnwidth]{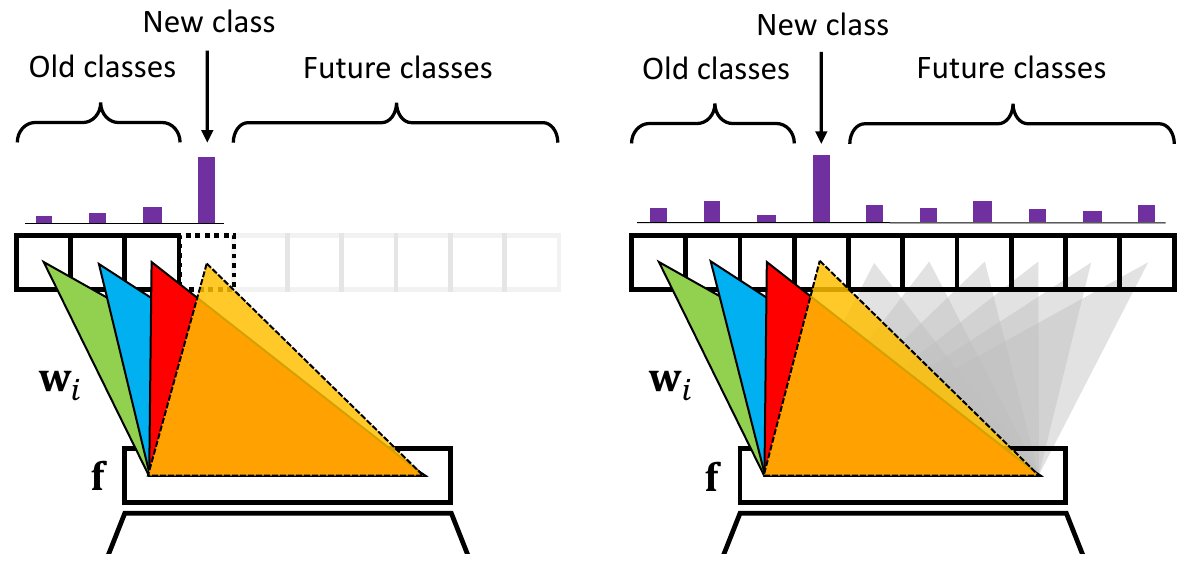}
    }
    \vspace{0.1cm} 
    \caption{ Class-incremental learning classifiers. Comparison between a standard \emph{expanding} classifier (left) and a \emph{pre-allocated} Regular Polytope Classifier (RPC) (right). Both classifiers are shown with three learned classes (green, blue, red) and with a new class under learning (orange). \emph{(a):} The weights of the classifiers are represented in feature space $\mathbf{f}$. The figure illustrates the situation in a 2D scenario, the characterization extends to arbitrary dimensions.
    As a new class is learned  old classes move to include the new one, the motion is shown in transparent colors. This effect is not present in our RPC. \emph{(b):} The final layers of the classifiers in a Neural Network architecture. The pre-allocated RPC weights (grey) can receive negative samples and adapt the network from the beginning of learning as new classes arrive. The logit responses of each class are also shown in purple. }
    \label{fig:RPCvsSTD}
\end{figure}

\subsection{Class Incremental Learning Setting}
In continual learning, a stream of data triplets $(x_i, y_i, t_i)$ containing an input $x_i$, a target $y_i$, and a task identifier $t_i \in \mathcal{T} = \{1, \ldots, T\}$ are presented to the learning agent.
Each input-target pair $(x_i, y_i) \in \mathcal{X} \times \mathcal{Y}_{t_i}$ is an identically and independently distributed (i.i.d.) example drawn from an unknown distribution $P_{t_i}(X, Y)$ that represents the $t_i$-th learning task.
We assume that the tasks are learned in order: $t_i \leq t_j$ for all $i \leq j$, and that the number of tasks $T$ is not known a priori. Specifically in CIL: a single classifier is learned and the task-membership $t_i$ is ignored. 
Under this setup, the goal is to estimate a Neural Network based model
\begin{equation}
f_{\theta} = (\mathbf{w} \circ \Phi) : \mathcal{X} \times \mathcal{T} \to \mathcal{Y}, 
\label{eq_predictor}
\end{equation}
parameterized by $\theta \in \mathbb{R}^p$ where $p$ is the number of parameters of $f_\theta$. The Neural Network model is composed of a feature extractor $\Phi : \mathcal{X} \to \mathcal{H}$ and a classifier $\mathbf{w} : \mathcal{H} \to \mathcal{Y}$, that minimize the multi-task loss
\begin{equation}
\mathcal{L}=
    \frac{1}{T} \sum_{t=1}^T \mathbb{E}_{(x, y) \sim P_{t}}\left[\, \ell(f(x, t), y) \,\right],
    \label{eq:multitask}
\end{equation}
where $\mathcal{Y} = \cup_{t \in \mathcal{T}} \mathcal{Y}_{t}$, and $\ell : \mathcal{Y} \times \mathcal{Y} \to \mathbb{R}$ is a loss function.

Following~\cite{lopez2017gradient}, we evaluate the performance of class-incremental learning algorithms according to the \emph{final average accuracy}
defined as
\begin{equation}
    \text{Accuracy} = \frac{1}{T} \sum_{j=1}^T a_{T, j},
    \label{eq:accuracy}
\end{equation}
where $a_{i,j}$ denotes the test accuracy on task $j$ after the model has finished learning the task $i$.
That is, the final average accuracy measures the test performance of the model at every task after the continual learning experience has finished.

\subsection{Class-Incremental Learning with a Pre-allocated Regular Polytope Classifier}

As based on Experience Replay (ER), our method learns   
the model $f_\theta$ by storing few past observed triplets in an episodic memory $\mathcal{M} = \{(x', y', t')\}$. For every new mini-batch of observations $\mathcal{B} = \{(x, y, t)\}$, the learner samples a mini-batch $\mathcal{B}_\mathcal{M}$ from $\mathcal{M}$ at random, and applies the rule 
$$
\theta \leftarrow \theta - \alpha \cdot \nabla_\theta \, \ell(\mathcal{B} \cup \mathcal{B}_\mathcal{M})
$$ to update the parameters of $f_\theta$, where
$$
    \ell(\mathcal{A}) = \frac{1}{|\mathcal{A}|} \sum_{(x, y, t) \in \mathcal{A}} \ell(f_\theta(x,t), y)
$$
denotes the average loss across a collection of triplets $\mathcal{A}$. 
Since we address class-incremental learning (i.e. single output head) the task identifiers $t$ and $t^\prime$ are ignored.

As firstly noted in \cite{rebuffi2017icarl}, in this learning condition, it is problematic that the classifier weight vectors $\mathbf{w}$ are decoupled from the feature extraction routine $\Phi$: whenever $\Phi$ changes in Eq.\ref{eq_predictor}, all $\mathbf{w}$ parameters must be updated as well. Otherwise, the network outputs will change uncontrollably, which is observable as catastrophic forgetting. 
Changes in the extracted features are mainly due to the inclusion of novel classes, that is, when a novel class is incorporated in the neural network model, the classifier weights of the other old classes move to create space to accommodate the novel one. Under the assumption of normalized weights and zero biases for the classifier, as proposed in  \cite{wang2017normface} and \cite{Liu2017CVPR}, the classifier weights are constrained in the unit hyper-sphere and can be easily visualized. Fig.~\ref{fig:RPCvsSTD}(a)(left) shows the geometric configuration of the classifier weights in feature space $\mathbf{f}$. 
As the new class (orange) is learned, old classes (green, blue, red) move changing their spatial configuration to include the new one. The motion is shown in transparent colors. Basically the set of weights forming an irregular triangle transforms into a set of weights forming an irregular quadrilateral and eventually, if a further class is introduced, the set of weights transforms into an irregular pentagon and so on. As new classes are continually incorporated into the model, the classifier continues to change its configuration (without stopping) with its corresponding features following a similar motion. The same transformative pattern occurs in higher dimensional spaces where sets of weights form convex polytopes.
Fig.~\ref{fig:RPCvsSTD}(b)(left) shows the corresponding expanding classifier in which the three old classes have been already learned (green, blue, red) and a new class is undergoing learning (orange).

In order to avoid this continuous motion of features, our approach uses a pre-allocated special fixed classifier (i.e. not undergoing the learning procedure) that keeps the features of the learned classes in a constant specific spatial configuration as novel classes are incorporated into the learning model. This allows to partially handle the catastrophic forgetting effect of the final classifier layer and to learn features that do not change their geometric configurations as  novel classes are incorporated in the  learning  model.  Fig.~\ref{fig:RPCvsSTD}(a)(right) shows the proposed fixed classifier consisting of a number of pre-allocated directions (grey) distributed at equal angles maximizing the available feature space with the purpose of defining class decision regions (delimited by decision boundaries) of equal extension for each class. Fig.~\ref{fig:RPCvsSTD}(b)(right) shows the corresponding  pre-allocated fixed classifier.

The number of pre-allocated classes is typically large because the number of class $K$ is not known a priori, however, this design choice allows our method to receive and learn from negative examples since the beginning of the data stream. More specifically,  Fig.~\ref{fig:RPCvsSTD}(a)(right) and Fig.~\ref{fig:RPCvsSTD}(b)(right) show how this is achieved: at each learning update, the weights of the classifier keep the same constant position \cite{pernici2019fix}.
By fixing the weights, the trainable classifier is superseded by previous layers. Fixed classifiers are shown recently to cause little or no reduction in classification performance for common datasets while allowing a noticeable reduction in trainable parameters, especially when the number of classes is large \cite{hoffer2018fix,pernici2019fix}.

Since no prior assumption about the semantic similarity between future classes can be made, in order to define the fixed classifier the natural assumption is to use the $d$-Simplex regular polytope. With the $d$-simplex, \emph{all classes are nearest to all other}.   
In geometry, a simplex is a generalization of the notion of a triangle or tetrahedron to arbitrary dimensions. Specifically, a $d$-simplex is a $d$-dimensional polytope which is the convex hull of its $d + 1$ vertices. A regular $d$-simplex may be constructed from a regular $(d-1)$-simplex connecting a new vertex to all original vertices by the common edge length. According to this, the weights for this classifier can be computed as: 
\begin{equation}
\mathbf{W}_\mathcal{S}=\Big \{e_1,e_2,\dots,e_{d-1}, \alpha \sum_{i=1}^{d-1} e_i \Big \}
\nonumber
\end{equation}
where $\alpha=\frac{1-\sqrt{d+1}}{d}$ and $e_i$ with $i \in \{1,2, \dots, d-1\}$ denote the standard basis in $\mathbb{R}^{d-1}$. The final weights will be shifted about the centroid and normalized.
The $d$-Simplex fixed classifier defined in an embedding space of dimension $d$, can accommodate a number of classes $K$ equal to its number of vertices:
\begin{equation}
K=d+1.
\label{eq_ksimplex}
\end{equation}
This classifier has the largest number of classes that can be embedded in $\mathbb{R}^d$ such that their corresponding class features are equidistant from each other. 

\vspace{0.5cm}

\begin{table*}
\centering
\caption{Accuracy (Eq.~\ref{eq:accuracy}) of class incremental learning experiments. Averages and standard deviations are computed over 10 runs using different random seeds. }
    \begin{tabular}{l|rccc|rc}
    \toprule
    \multicolumn{1}{l}{\textbf{Method}} &
    \multicolumn{1}{|r}{{$~~|\mathcal{M}|$}} &
    \multicolumn{1}{c}{\textsc{SplitMNIST}} & 
    \multicolumn{1}{c}{\textsc{PermutedMNIST}} &
    \multicolumn{1}{c}{\textsc{SplitCIFAR10}} &
    \multicolumn{1}{|r}{{$~~|\mathcal{M}|$}} &
    \multicolumn{1}{c}{\textsc{SplitCIFAR100}}
    \\
    \midrule 
    EWC~\cite{Kirkpatrick2016EWC}       &    - & 19.92 $\pm$ 0.06 & 19.51 $\pm$ 0.05 & 16.89 $\pm$ 0.03 &    - & 17.17 $\pm$ 0.12\\
    Online-EWC~\cite{progresscompress}  &    - & 19.93 $\pm$ 0.05 & 31.63 $\pm$ 0.11 & 17.28 $\pm$ 0.09 &    - & 17.29 $\pm$ 0.06\\
    SI~\cite{zenke2017continual}        &    - & 21.01 $\pm$ 0.10 & 18.27 $\pm$ 0.07 & 17.81 $\pm$ 0.12 &    - & 14.21 $\pm$ 0.29\\
    MAS~\cite{aljundi2017memory}        &    - & 23.01 $\pm$ 0.31 & 17.53 $\pm$ 0.53 & 17.75 $\pm$ 0.87 &    - & 16.97 $\pm$ 0.05\\
    GEM~\cite{lopez2017gradient}        &  100 & 74.92 $\pm$ 2.97 & 31.03 $\pm$ 3.19 & 24.48 $\pm$ 3.21 & 1400 & 22.83 $\pm$ 2.17\\
    GEM~\cite{lopez2017gradient}        & 1100 & 95.16 $\pm$ 0.15 & 79.44 $\pm$ 0.23 & 45.48 $\pm$ 0.19 & 5600 & N.A.\\
    Expanding Classifer              &  100 &
        80.10 $\pm$ 3.29 &
        64.18 $\pm$ 2.85 & 
        31.74 $\pm$ 2.15 &
        1400 &
        33.79 $\pm$ 0.56\\
    Expanding Classifer              & 1100 &
        96.25 $\pm$ 0.28 &
        93.52 $\pm$ 0.41 &
        66.93 $\pm$ 0.48 &
        5600 &
        51.76 $\pm$ 0.55\\
    \midrule
    Pre-allocated RPC (Ours)            &  100 &
        82.32 $\pm$ 2.19 &
        81.43 $\pm$ 2.12 &
        31.80 $\pm$ 1.61 & 
        1400 &
        33.80 $\pm$ 0.42 \\
    Pre-allocated RPC (Ours)            & 1100 & 
        96.90 $\pm$ 0.29 &
        94.63 $\pm$ 0.35 &
        67.44 $\pm$ 0.50 &
        5600 &
        51.77 $\pm$ 0.61 \\
        
    \bottomrule
    \end{tabular}%
\label{table:mnist-cifar}
\end{table*}

\subfigcapskip = 0.1cm
\subfigbottomskip = 0.3cm
\begin{figure*}
    \subfigure[SplitMNIST]{
        \includegraphics[width=0.66\columnwidth]{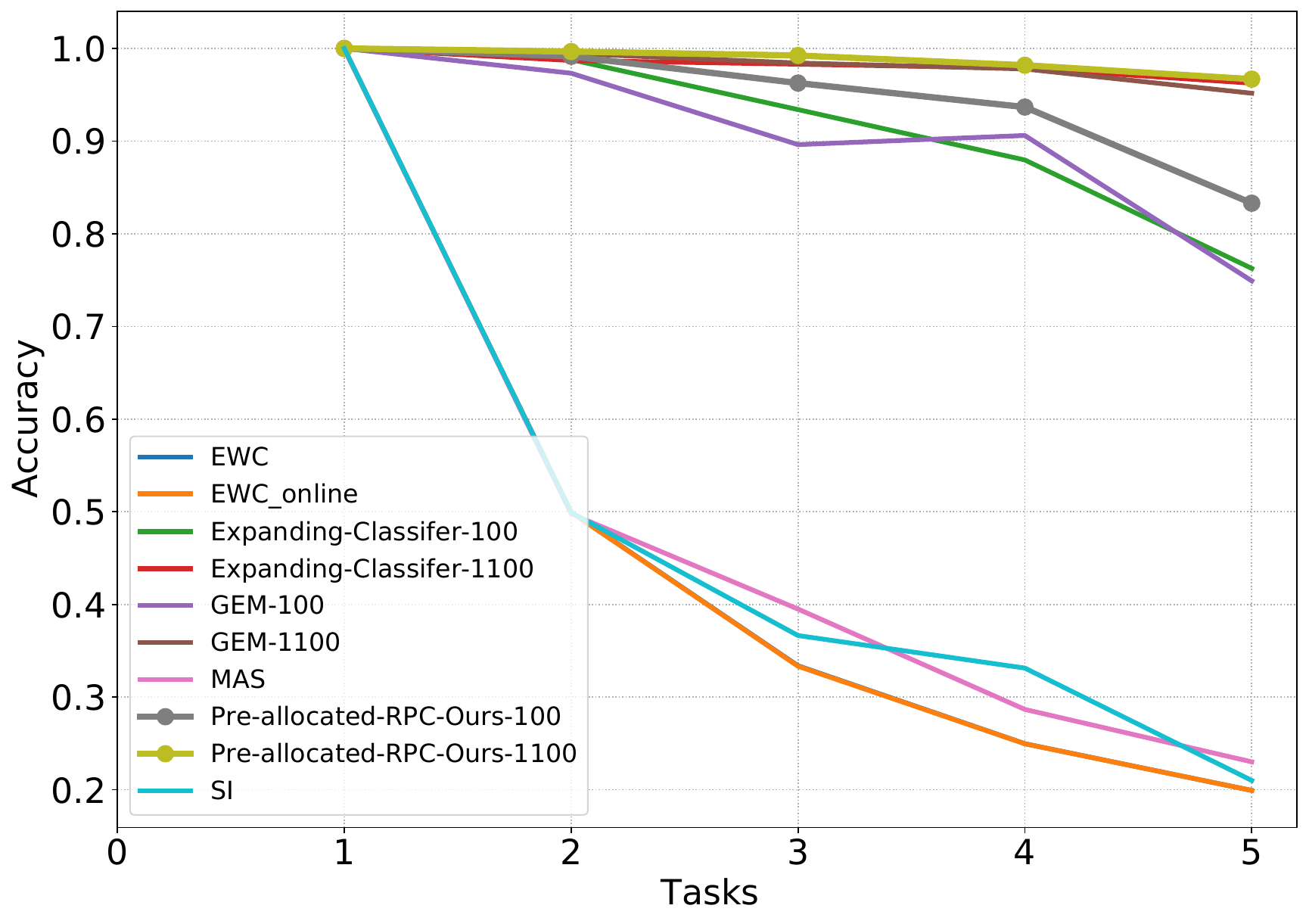}
        \label{fig:evolution-splitmnist}
    }
    \subfigure[PermutedMNIST]{
        \includegraphics[width=0.66\columnwidth]{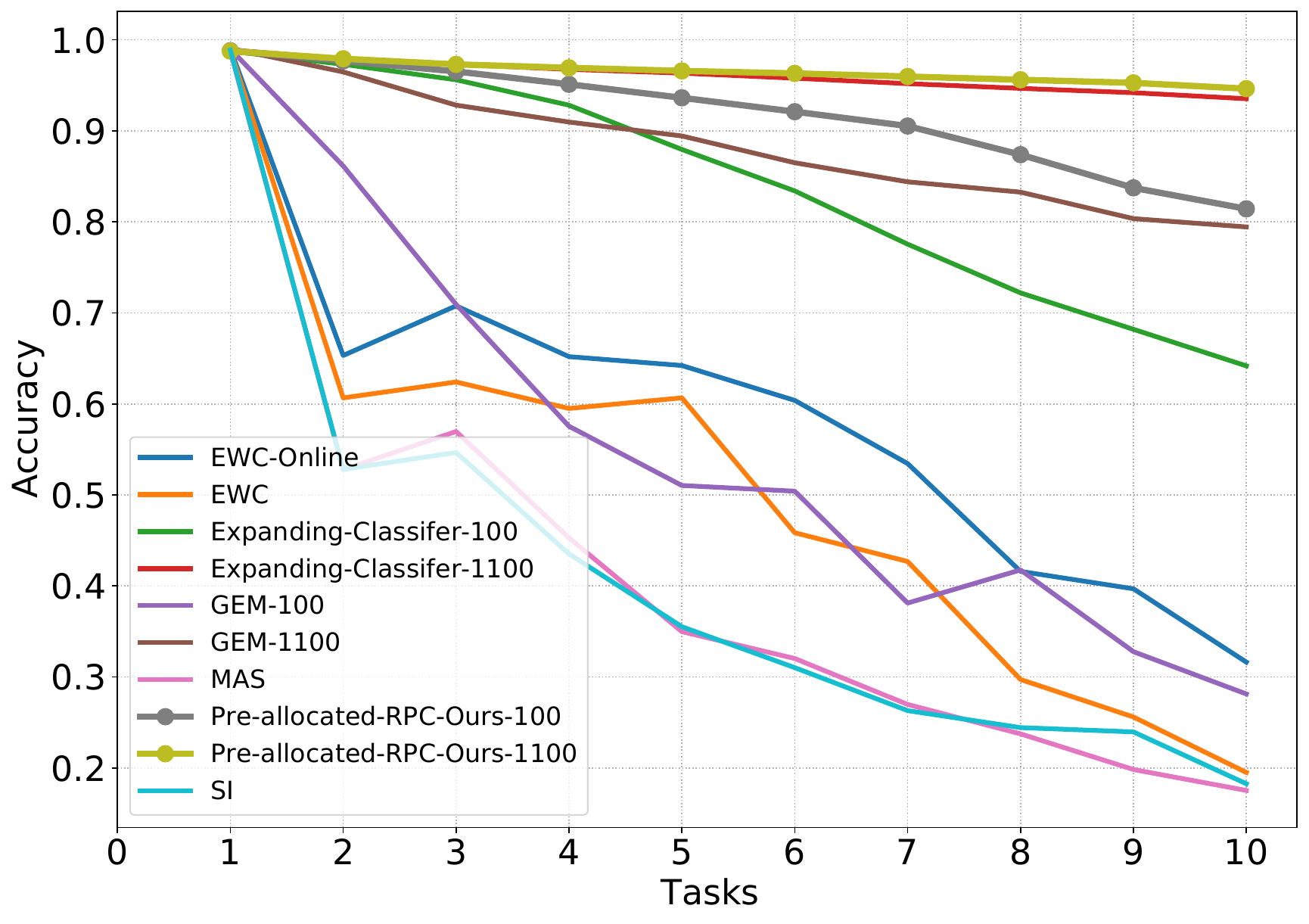}
        \label{fig:evolution-permutedmnist}
    }
    \subfigure[SplitCIFAR10]{
        \includegraphics[width=0.66\columnwidth]{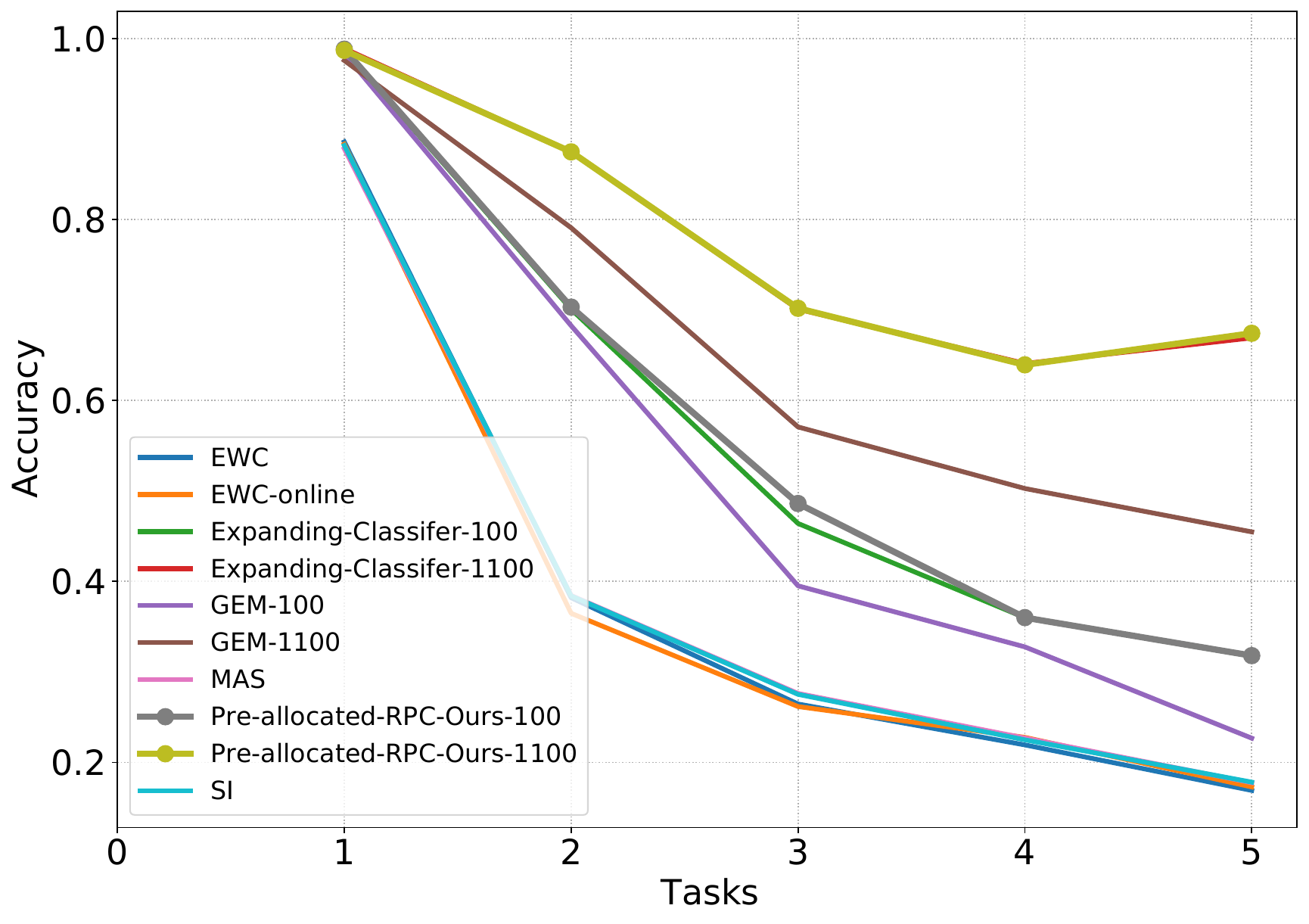}
        \label{fig:evolution-cifar10}
    }
    \caption{Evolution of accuracy as new tasks are learned. \emph{(a)}, \emph{(b)} and \emph{(c)} show the results for SplitMNIST, PermutedMNIST and SplitCIFAR10, respectively.
    \vspace{0.4cm}}
    \label{fig:evolution}
\end{figure*}

\section{Experimental Results}

We perform experiments on four classiﬁcation benchmarks for continual learning: SplitMNIST, PermutedMNIST, SplitCIFAR10 and SplitCIFAR100. SplitMNIST splits the MNIST dataset of handwritten digits \cite{lecun1998mnist} to create five different tasks with non-overlapping classes. PermutedMNIST is a variant of the MNIST dataset where each task applies a ﬁxed random pixel permutation to the original. 
A total of ten tasks of ten classes each is created.
{Although this dataset is unrealistic from the point of view of how images are formed \cite{farquhar2018towards}, PermutedMNIST allows evaluating and understanding class-incremental learning systems in the extreme case in which tasks are unrelated to each other \cite{Hsu18_EvalCL,wortsman2020supermasks}. }
SplitCIFAR10 and SplitCIFAR100 are variants of the CIFAR10 and CIFAR100 datasets, respectively \cite{krizhevsky2009learning,zenke2017continual}. 
In CIFAR100 each task contains the data pertaining to twenty random classes out of the total 100 classes. This benchmark contains five tasks. In CIFAR10 a total of five tasks of two classes each are created.
 
We implemented our fixed classifier on top of the LeNet architecture \cite{lecun1998gradient} for the MNIST and PermutedMNIST datasets. %
Popular network architectures for ImageNet require modifications to adapt to the CIFAR 32x32 input size. According to this, for the SplitCIFAR10 and SplitCIFAR100 dataset our experiments follow publicly available implementations\footnote{
\url{https://github.com/GT-RIPL/Continual-Learning-Benchmark/} } and for our fixed classifier implemented on top of a {ResNet56} architecture.

We compared our proposed model to the following baselines: EWC \cite{kirkpatrick2017overcoming}, Online EWC \cite{progresscompress}, SI \cite{zenke2017continual}, MAS \cite{aljundi2017memory}, GEM \cite{lopez2017gradient} and Expanding Classifier with Experience Replay.
For both the Expanding Classifier and our method, the mini-batch is constructed by an equal amount (64/64) of new data and the memory data. The buffer size is predefined to match the space overhead used by Online-EWC and SI, which translates to 1100 and 5400 images for the MNIST/CIFAR10 and CIFAR100 datasets, respectively \cite{Hsu18_EvalCL}.
\begin{figure}[!b]
    \centering 
    \includegraphics[width=.77\columnwidth]{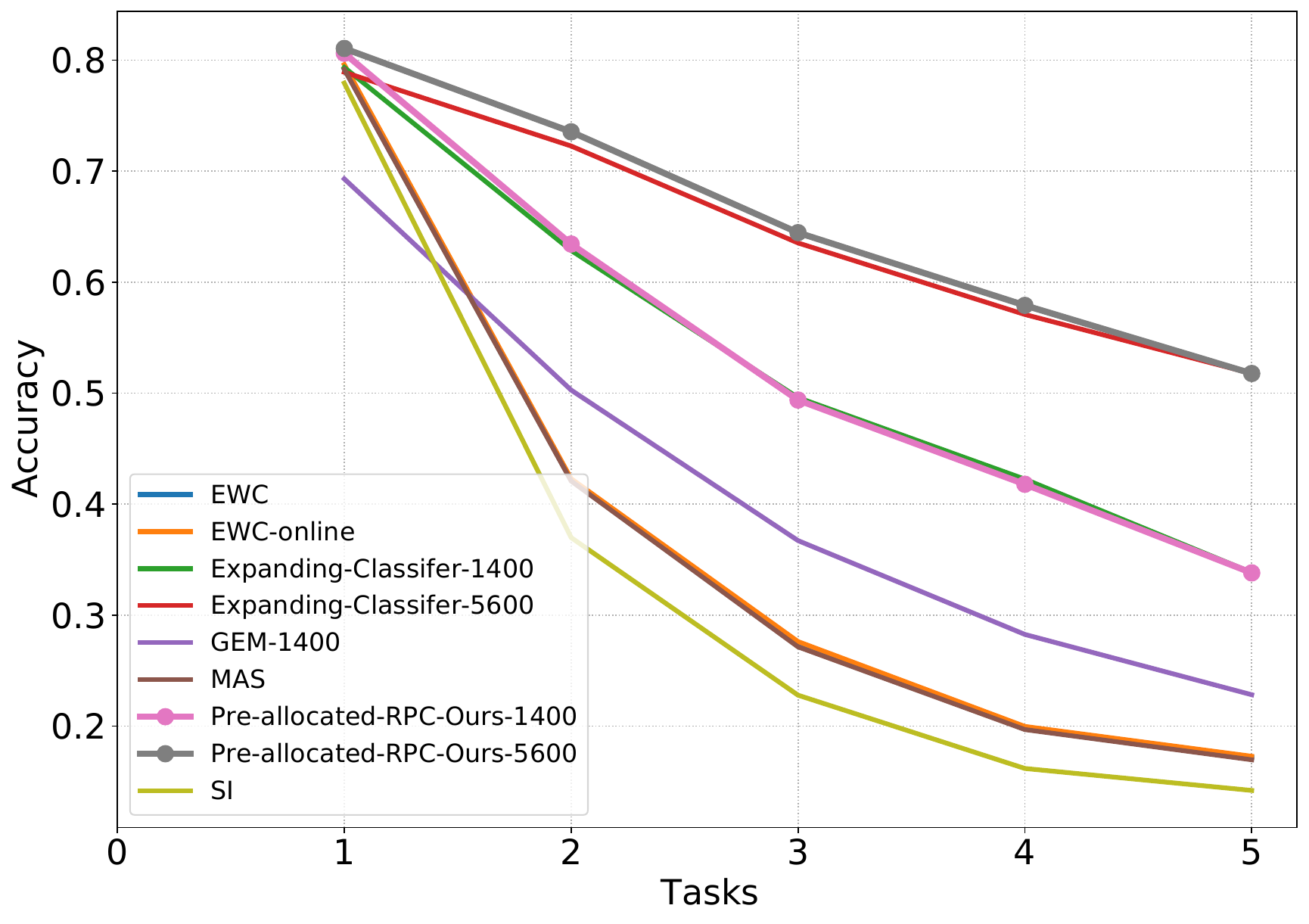}
    \caption{Evolution of Accuracy for the SplitCIFAR100 dataset.}
    \label{fig:evolutioncifar100}
\end{figure}
For a fair comparison, all methods use the same neural network architecture. 
The classification loss function is the standard cross-entropy in all methods. All models are trained for 5 epochs per task with mini-batch size 128 using the Adam optimizer ($\beta_1 = 0.9$, $\beta_2 = 0.999$, learning rate = 0.001) as the default, unless explicitly described. SplitCIFAR10 and SplitCIFAR100 are trained with 10 epoch per task.
For reproducibility, all the results of the baselines are evaluated from scratch (no results are reported from papers) based on the unified code described in \cite{Hsu18_EvalCL}.

\begin{table*}
\centering
\caption{ Ablation study on CIFAR-10 for different number of pre-allocated classes.
 }
    \begin{tabular}{lrcccc}
    \toprule
    & & \multicolumn{4}{c}{\textsc{SplitCIFAR10}} \\
    & & \multicolumn{4}{c}{\textsc{\#Pre-allocated Classes}} \\
    \cmidrule(lr){3-6}
    \multicolumn{1}{l}{\textbf{Method}}
    &\multicolumn{1}{r}{{$|\mathcal{M}|$}}
    &\multicolumn{1}{c}{{10}} & \multicolumn{1}{c}{{50}}
    &\multicolumn{1}{c}{{100}} &\multicolumn{1}{c}{{1000}} \\
    \midrule 
    Pre-allocated Expanding Classifer             &    100 & 
        31.74 $\pm$ 2.15  &
        31.59 $\pm$ 0.96  &
        31.64 $\pm$ 1.37  &
        32.05 $\pm$ 1.10  \\
    Pre-allocated Expanding Classifer              &    1100 &
        66.93 $\pm$ 0.48  &
        66.57 $\pm$ 0.90  &
        66.37 $\pm$ 1.20  &
        66.75 $\pm$ 0.68  \\
    \midrule
    Pre-allocated RPC (Ours)            & 100 & 
        31.80 $\pm$ 1.61 &
        32.20 $\pm$ 0.70 &
        32.93 $\pm$ 1.50 &
        32.27 $\pm$ 1.31 \\
    Pre-allocated RPC (Ours)            & 1100 & 
        67.44 $\pm$ 0.50 &
        67.07 $\pm$ 0.12 &
        67.10 $\pm$ 0.26 & 
        66.77 $\pm$ 0.52 \\
        
    \bottomrule
    \end{tabular}%
\label{table:cifar-10}

\end{table*}

\begin{table*}
\centering
\caption{ Ablation study on CIFAR-100 for different number of pre-allocated classes. }
    \begin{tabular}{lrcccc}
    \toprule
    & & \multicolumn{4}{c}{\textsc{SplitCIFAR100}} \\
    & & \multicolumn{4}{c}{\textsc{\#Pre-allocated Classes}} \\
    \cmidrule(lr){3-6}
    \multicolumn{1}{l}{\textbf{Method}}
    &\multicolumn{1}{r}{{$|\mathcal{M}|$}}
    &\multicolumn{1}{c}{{100}} & \multicolumn{1}{c}{{200}}
    &\multicolumn{1}{c}{{500}} &\multicolumn{1}{c}{{1000}} \\
    \midrule 
    Pre-allocated Expanding Classifer               &    1400 &
        33.79 $\pm$ 0.56  &
        33.07 $\pm$ 0.61  &
        33.37 $\pm$ 0.33  &
        33.55 $\pm$ 0.48 \\
    Pre-allocated Expanding Classifer              &    5600 &
        51.76 $\pm$ 0.55  &
        49.71 $\pm$ 0.63  &
        49.18 $\pm$ 0.85  &
        48.62 $\pm$ 0.96 \\
    \midrule
    Pre-allocated RPC (Ours)            & 1400 & 
        33.80 $\pm$ 0.42 &
        33.43 $\pm$ 0.45 &
        33.98 $\pm$ 0.38 & 
        33.92 $\pm$ 0.47 \\
    Pre-allocated RPC (Ours)            & 5600 & 
        51.77 $\pm$ 0.61 &
        51.25 $\pm$ 0.39 &
        51.28 $\pm$ 0.49 &
        51.39 $\pm$ 0.52 \\
        
    \bottomrule
    \end{tabular}%
\label{table:cifar-100}

\end{table*}

Tab.~\ref{table:mnist-cifar} summarizes the main results of our experiments. An ablation study of memory size is also included (100 and 1100 elements for SplitMNIST, PermutedMNIST and SplitCIFAR10; 1400 and 5600 elements for CIFAR100).
First, our proposed method achieves similar accuracy (Eq.\ref{eq:accuracy}) to the expanding classifier in all benchmarks. Second, the relative gains from the same methods using one order small memory, namely GEM and Expanding Classifier with 100 memory elements are signiﬁcant, conﬁrming that the pre-allocated fixed classifier allows Experience Replay methods to work better with less memory. Third, note that approaches making use of memory (GEM, ER based methods) work signiﬁcantly better in this setup, while regularization methods such as EWC, Online-EWC, SI and MAS are suffering the class-incremental learning setting. {Note that for GEM with 5600 memory elements evaluated in the SplitCIFAR100 dataset did not complete due to lack of GPU memory resources.}

Fig.~\ref{fig:evolution} and Fig.~\ref{fig:evolutioncifar100} show a more detailed  analysis of the average accuracy as new tasks are incrementally learned. More specifically, Fig.~\ref{fig:evolution}(a), (b), (c) and Fig.~\ref{fig:evolutioncifar100} show SplitMNIST, PermutedMNIST, SplitCIFAR10, and SplitCIFAR100, respectively. As evident from all the figures the performance of our approach is no worse than other baselines and in some cases, the accuracy is slightly higher. In the case of PermutedMNIST with 100 memory elements is clearly higher.

\begin{figure*}
    \centering
    \subfigure[]{
    \includegraphics[trim=0 8cm 0 0,clip,width=0.95\textwidth]{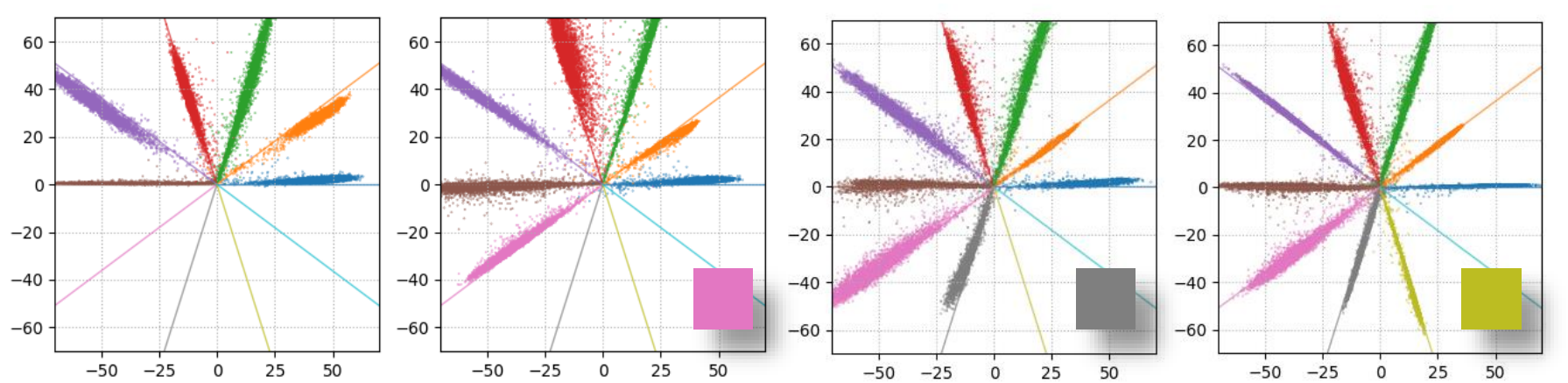}
    }
    \subfigure[]{
    \includegraphics[trim=0 0 0 8cm,clip,width=0.95\textwidth]{figures/continual__Features.pdf}
    }
    \caption{Class Incremental Feature Learning on the MNIST dataset in a 2D embedding space. Figures (a) and (b) show the 2D features learned by the RPC classifier and by a standard trainable classifier, respectively. Learning starts from six classes and three more are added, one at a time (colored squares), with ten pre-allocated classes. The two methods achieve the same classification accuracy. The figures show the training evolution of the classifier weights (colored lines) and their corresponding test-set class feature (2D point cloud), respectively. As it can be noticed, in (b) each novel learned class significantly and unpredictably changes the geometric configuration of already learned features (major changes highlighted by circles). As shown in (a) this effect is absent in the RPC classifier. The figure is best viewed in color.}
    \label{fig:feature_evolution}
\end{figure*}

{\bf{Ablation Study}.}
We conducted an ablation study on the effect of pre-allocation on both the RPC classifier and the Expanding Classifier. Different number of pre-allocated classes are also evaluated. The quantitative results in Tab.~\ref{table:cifar-10} and Tab.~\ref{table:cifar-100} for the CIFAR10 and CIFAR100 datasets, respectively, show that in both cases the pre-allocation does not substantially affect the final performance. The advantage of our method is that the geometric configuration of features in our method does not change as novel classes are incorporated into the CNN model. This property can be appreciated in  Fig.~\ref{fig:feature_evolution}. In particular, Fig.~\ref{fig:feature_evolution}(a) shows the evolution of the distribution of class features as learned by a $10$-sided polygon fixed classifier and Fig.~\ref{fig:feature_evolution}(b) shows the evolution of a standard pre-allocated Expanding Classifier. Both classifiers are implemented on top of the LeNet architecture and learned using the MNIST dataset. This toy example reduces the output size of the last hidden layer to 2 (i.e. the dimension of the feature is 2) so that we can directly plot the distribution of features on a 2D plane to ease visualization. A $k$-sided polygon is the equivalent of a Regular Polytope Classifier in a two dimensional space. While the figure illustrates the situation in $\mathbb{R}^2$, the characterization extends to arbitrary dimensions. The evolution of the incrementally learned features is shown starting from three classes and adding two more, one at a time, with ten pre-allocated classes. In the case of the pre-allocated Expanding Classifier, weights are randomly initialized. As it can be noticed in Fig.~\ref{fig:feature_evolution}(b), each novel learned class, significantly and unpredictably changes the geometric configuration of the already learned features and already learned classifier weights. This effect is particularly evident in the highlighted classes. Contrarily, as shown in Fig.~\ref{fig:feature_evolution}(a), our method is not influenced by this effects: both features (colored point cloud) and weights (colored lines) do not change their relative geometric configuration. In addition to this, class feature distributions are perfectly aligned with their corresponding class weights.

\vspace{0.5cm}
\section{Conclusion}
We introduced a novel approach for class-incremental learning that exploits future unseen classes as negative examples and learns features that do not change their geometric configuration as novel classes are  incorporated in the learning model. The approach uses a pre-allocated special fixed classifier (i.e. not undergoing the learning procedure) in which weights are set according to the vertices of the $d$-Simplex regular polytope. As shown in the experiments our method is as effective as the expanding classifier while exhibiting properties of internal feature representation that are otherwise not-existent.
Our finding may have implications in those Deep Neural Network incremental learning contexts in which feature representation must be compatible with previously learned representations.

\section*{Acknowledgement} This work was partially supported by the Italian MIUR within PRIN 2017, Project Grant 20172BH297: I-MALL. 

\bibliographystyle{IEEEtran}
{\Large
\bibliography{IEEEexample}
}

\end{document}